\pdfoutput=1 

\documentclass[letterpaper,twocolumn,fleqn]{article} 

\usepackage{ist}
\usepackage{times}

\usepackage{soul}
\usepackage{url}
\usepackage[hidelinks]{hyperref}
\usepackage[utf8]{inputenc}
\usepackage[small]{caption}
\usepackage{graphicx}
\usepackage{amsmath}
\usepackage{booktabs}
\usepackage[symbol]{footmisc}

\usepackage{subfig}
\usepackage{xcolor}

\title{Improving Food Detection For Images \\From a Wearable Egocentric Camera}

\author{ 
Yue Han; School Of Electrical and Computer Engineering, Purdue University; West Lafayette, IN\\
Sri Kalyan Yarlagadda; School Of Electrical and Computer Engineering, Purdue University; West Lafayette, IN\\
Tonmoy Ghosh; Department of Electrical and Computer Engineering, The University of Alabama; Tuscaloosa, AL\\
Fengqing Zhu; School Of Electrical and Computer Engineering, Purdue University; West Lafayette, IN\\
Edward Sazonov; Department of Electrical and Computer Engineering, The University of Alabama; Tuscaloosa, AL\\
Edward J. Delp; School Of Electrical and Computer Engineering, Purdue University; West Lafayette, IN\\
}

\date{} % date has an empty field.

\begin{document} 

\maketitle 

\footnote[0]{Research reported in this publication was partially by the National Institute of Diabetes and Digestive and Kidney Diseases (grants number: R01DK100796) and by the endowment of the Charles William Harrison Distinguished Professorship. The content is solely the responsibility of the authors and does not necessarily represent the official views of the National Institutes of Health.}

\thispagestyle{empty} % prevents the first page to be numbered

%%%%%%%%%%%%%%%%%%%%%%%%%%%%%%%%%%
% Abstract
%%%%%%%%%%%%%%%%%%%%%%%%%%%%%%%%%%

\begin{abstract}
Diet is an important aspect of our health. Good dietary habits can contribute to the prevention of many diseases and improve overall quality of life. To better understand the relationship between  diet and health,   image-based dietary assessment systems have been developed to collect dietary information. 
%Some systems require users to use a mobile phone to capture images of their eating scenes.  While mobile phones are ubiquitous and easy to use, they still require manual effort from the users. 
We introduce the Automatic Ingestion Monitor (AIM), a device that can be attached to one's eye glasses. It provides an automated hands-free approach to capture eating scene images. While AIM has several advantages, images captured by the AIM are sometimes blurry. Blurry images can significantly degrade the performance of food image analysis such as food detection. In this paper, we propose an approach to pre-process images collected by the AIM imaging sensor by rejecting extremely blurry images to improve the performance of food detection.
\end{abstract}

%%%%%%%%%%%%%%%%%%%%%%%%%%%%%%%%%%%%
% Overall Document Guidelines: Head
%%%%%%%%%%%%%%%%%%%%%%%%%%%%%%%%%%%%
\section{Introduction}
\label{sec:intro}

In 2016, \$7.5 trillion was spent on healthcare worldwide, which is approximately 10\% of the world GDP~\cite{world2018public}. At the same time, over \$50 billion per year was spent on diet-related cardiometabolic disease~\cite{CMDcost}. Understanding the factors that influence  health can help prevent this unnecessary expenditure and associated illness.  It is well-known that  dietary habits have a profound impact on  health~\cite{Mesas2012}. Poor dietary habits can contribute to ailments such as heart diseases, diabetes, cancer, and obesity. Dietary studies have shown that an unhealthy diet, such as skipping meals, can also be linked to stress, depression, and other mental illness~\cite{2}. Because poor diet could have such a severe impact on our health, it is important that we  study and understand the complex relationship between  dietary habits and health.

To achieve this goal, nutrition practitioners and dietary researchers conduct dietary studies to collect data about the dietary habits of people. The collected data is used to analyzed and to understand how dietary patterns influence  health. A challenging aspect of conducting these dietary studies is the data collection process. Self-reporting techniques such as Food Frequency Questionnaire (FFQ) and Automated Self-Administered Recall System (ASA24) are standard tools used to collect such data~\cite{Dietmethod}. The accuracy of the data depends on the participants' motivation and the ability to accurately remember their diet. In addition, they can be time consuming and laborious. To overcome these difficulties  several techniques to automatically collect dietary data have been developed. 
Some rely on images of eating scenes to extract dietary information. These techniques are referred to as image-based dietary assessment methods. TADA~\cite{Zhu2010}, FoodLog~\cite{food_log}, DietCam~\cite{diet_cam}, and FoodCam~\cite{8} are examples of image-based dietary assessment methods. All these systems require users to take a picture of the eating scenes using mobile telephones. These eating scene images are then analyzed by trained dietitians to estimate the nutrient information. This process is also time consuming, costly and laborious. Recently, progress has been made on automating this process~\cite{fang-icip2018,fang-globalsip2017,fangism2015,food_portion1,fp2,extra1,extra2,extra3}. The process of extracting nutrient content in an image involves 3 sub-tasks, food detection and segmentation, food classification and portion size estimation~\cite{Zhu2010,carkorie3task,smartlog,extra4}.

Mobile telephones are ubiquitous in today's society and used by all age groups of the population. Using mobile telephones for image-based dietary assessment makes the process simple, cheap, and easy to use. 
However, taking out the mobile telephones during eating requires manual effort and may be inconvenient. 
To overcome this challenge, we will describe the Automatic Ingestion Monitor (AIM) which provides a hands-free approach to automatically capture food images during an eating occasion.

AIM is a passive food intake sensor that requires no self-reporting during the eating occasion and can be easily mounted on eyeglasses. Additionally, this device also automated the entire image extraction process. Food intake is detected by the built-in accelerometer{~\cite{AIM1_2012, AIM1_2014, AIM2_2016, farooq2018accelerometer, AIM2_2020}}. %AIM is also capable of precisely estimating the chew count during an eating occasion.
%\textbf{{\color{red} You need to put some reference cites in here to AIMS}}
The images are stored on a SD card and can be exported by USB interface or Bluetooth. While AIM automatically captures eating scene images, these images are sometimes affected by motion blur. Blurry images can potentially reduce the performance of image analysis such as food detection, food segmentation, food classification and portion size estimation.  In this paper, we propose a method to automatically detect and remove extremely blurry images from the training dataset to improve the accuracy of food detection.

\section{Automatic Ingestion Monitor (AIM)}
\label{sec:aim}
\begin{figure}[t]
    \centering
    \includegraphics[scale = 0.6]{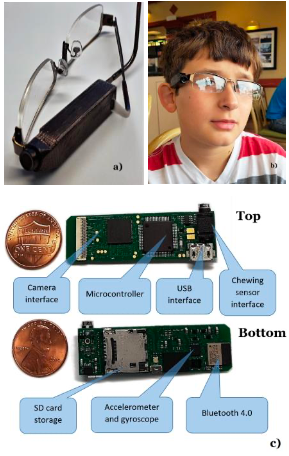}
    \caption{\textbf{a)} AIM mounted on eyeglasses. \textbf{b)} A child wearing AIM. \textbf{c)} Electronics of AIM.}
    \label{fig:aim}
\end{figure}

AIM is a device that clips on to eye glasses. It consists of the following sensors (see Figure~\ref{fig:aim})
\begin{itemize}
    \item 5 Megapixel camera
    \item Accelerometer 
    \item Curved strain sensor
\end{itemize}

%Please refer to Figure~\ref{fig:aim} for a visual description of AIM. 
The camera sensor is aligned with the person's eye gaze and is used to capture images of the eating scene. The accelerometer is used for food intake detection. 
The curved strain sensor is in contact with the temporalis muscle and provides a precise estimate of the chew count. Images are captured periodically every 15s. The onboard SD card has a capacity of storing images captured continuously for more than 4 weeks.
 Recent community studies demonstrated that AIM is able to detect food-intake with an accuracy (F1 score) of $96\%$~\cite{farooq2018accelerometer}. 
 In addition, AIM's chew count estimate has a very low mean absolute error of $3.8\%$. 
% All the above characteristics of AIM make it a better alternative to mobile phones. Capturing eating scene images using mobile phones requires manual effort and sometimes can be uncomfortable. Mobile phones also don't give any information about the chew count. 
Chewing and swallowing are directly related to food-intake and chew count data can serve as estimators of ingested mass and energy intake~\cite{amft2009bite,fontana2015energy,yang2019}. AIM is also safe to use for regular food intake study, it is based on low power, low-voltage (3V) and poses no more than minimal risk, comparable to a consumer electronic device.

\section{Dataset Description}
\label{sec:dataset}
The image dataset was collected from thirty volunteers using AIM. 
It contained 20 males and 10 females, mean $\pm$ SD age of 23.5 $\pm$ 4.9 years, range 18-39 years, mean body mass index (BMI) 23.08 $\pm$ 3.11 $kg/m^2$, range 17.6 to 30.5 $kg/m^2$. 
The University of Alabama's Institutional Review Board (IRB) approved the study. 
Each volunteer wore AIM for 2 days, the second day being the free-living day with no restrictions being imposed on food intake or other activities. 
A total of 90170 images were captured by the AIM device during the free-living day with
5418 images were captured when AIM detected a food-intake session. 
Our dataset comprises these 5418 images. We manually labeled the  foods and beverages in  them via bounding boxes. The dataset is randomly split into 3 subsets namely training (4,333 images), validation (585 images), and testing (500 images). We report the frequency of appearance of different objects in each of the subsets of the dataset in Table~\ref{tab:obj_freq}.

\begin{table}
\centering
\begin{tabular}{llll}
\toprule
 Number of objects & Training & Validation & Testing \\
\midrule
food       & 4,033  & 570 & 472     \\
beverage   & 2,239  & 288 & 264   \\
\bottomrule
\end{tabular}
\caption{Number of food, beverage objects in Training, Validation, and Testing subsets}
\label{tab:obj_freq}
\end{table}

\section{Blur Detection}
\label{sec:blur}
\begin{figure}[t]
    \centering
    \includegraphics[scale = 0.28]{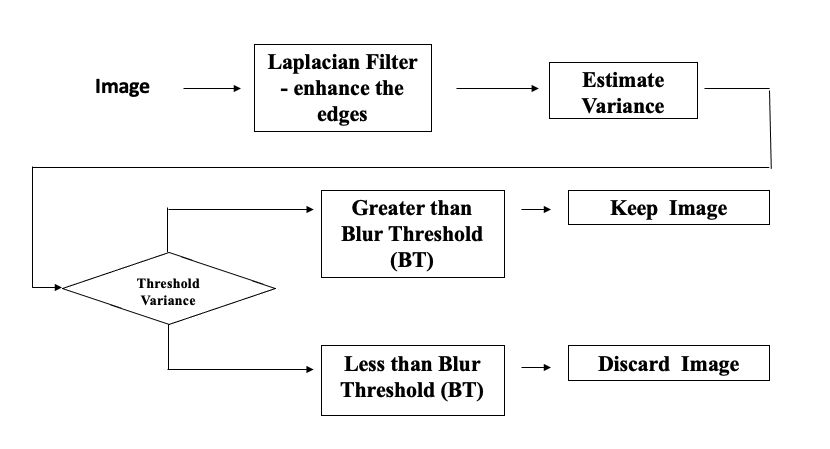}
    \caption{Block diagram of our blur detection method}
    \label{fig:blur_bd}
\end{figure}

\begin{figure}[t!]
	%\centering
	\subfloat[variance \textgreater \ BT]{\includegraphics[width =4cm]{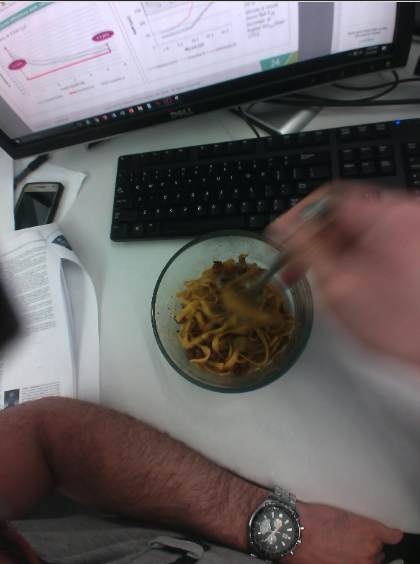}\label{fig:not_blurry}}
	\hfill
	\subfloat[variance \textless \  BT]{\includegraphics[width=4cm]{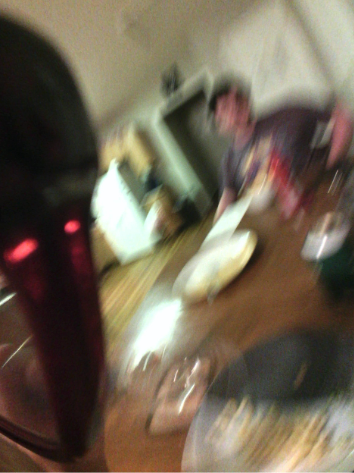}\label{fig:blurry}}
	\caption{Examples of images with variance less than the Blur Threshold (BT) and greater than BT. $BT = 10$.}
	\label{fig:examples}
\end{figure}

Before we describe our method, we will briefly describe how blurring occurs in images. Let $Y$ denote a blurry image and $I$ its non blurry counterpart then $Y$ and $I$ are related by Equation~\ref{eq:eq1}  
\begin{equation} \label{eq:eq1}
    B = b * I + W 
\end{equation}
Here $b$ is the blur kernel and $W$ is white Gaussian noise. $*$ denotes the convolution operation. The blur kernel is a low pass filter that suppresses the high frequency information in an image. 
The extent of loss of information depends on the frequency characteristics of $b$. This loss of high frequency information can be detected visually by inspecting the edge characteristics in the image. 
In blurry images, edges are hard to detect and extremely blurry images have no relevant object features. 
When designing a learning based food detection method, the presence of extremely blurry images in the training set could hamper the performance of image analysis.

Our blur detection process is summarized in Figure~\ref{fig:blur_bd}. 
We first estimate the blur in the image using an approach proposed in ~\cite{laplacian}. Blur in an image is estimated by using the Laplacian operator on the image and then estimating the variance of its output. 
The Laplacian operator has characteristics similar to a high pass filter and hence amplifies edge pixels in an image. If the variance is low, then it is likely that the image has blurry or "unsharp" edges. 
Thresholding the variance can be used to decide if an image is  blurred or not. We refer to this threshold as the Blur Threshold (BT). The selection of BT will be selected experimentally and will be discussed in the food detection section.  If the variance is less than BT, then the image is discarded from further image analysis. 
Figure~\ref{fig:blurry} shows an example image with a variance less than BT and Figure~\ref{fig:not_blurry} shows an image with a variance greater than BT. It's obvious that there is no relevant object (food/beverage) features in Figure~\ref{fig:blurry}. Figure~\ref{fig:not_blurry} has some blurry regions because of hand motion but regions belonging to objects of food are still clear enough for image analysis.

\section{Food Image Analysis}
\label{sec:analysis}
\begin{figure}[t]
    \centering
    \includegraphics[scale = 0.258]{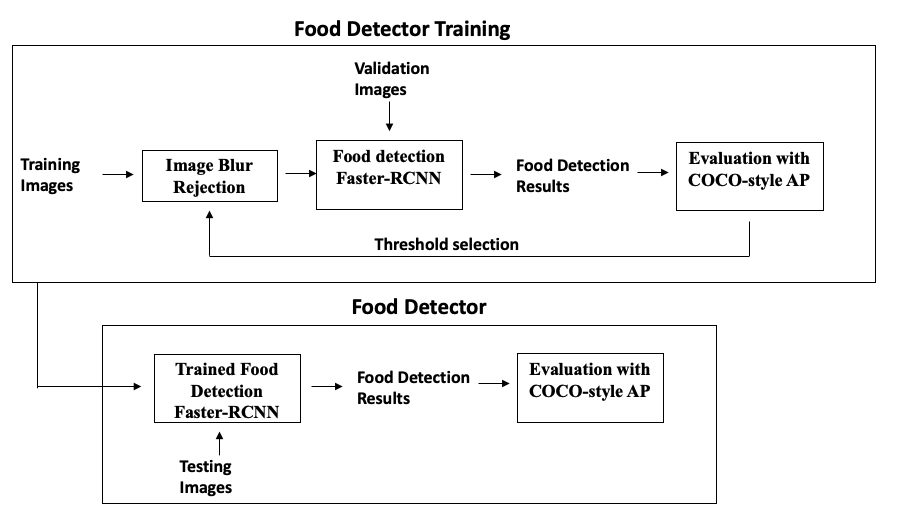}
    \caption{The block diagram of our food detection system}
    \label{fig:system_bd}
\end{figure}

Visual food-related information from eating occasions plays an important role in automatic dietary analysis. 
In this paper, we describe a "food/no food" detection task, detecting whether an AIM captured image has food/beverage present using the Faster-RCNN network~\cite{9}. 
By training on  blurry images, we are forcing the model to learn from data that has no relevant features in them. 
We used our blur detection technique described above to reject  blurry images from the training set. 
Our food detection system is summarized in Figure~\ref{fig:system_bd}. The training set images will first go through  image blur rejection  with several pre-selected thresholds. 
Then the images not rejected will be used as training data for the Faster-RCNN network for food detection. 
The system will be tested on the validation set and evaluated using the  COCO-style Average Precision(AP)~\cite{COCO}. The selected threshold, BT,  will be based on the performance of the food detection system on the validation set. 
We then choose the system with the selected BT threshold as the food detector used to detect the food on the testing  set,
Finally, we evaluate the results using the  COCO-style AP~\cite{COCO}.

\section{Experiments}
\label{sec:expriment}

\begin{table}
\centering
\resizebox{\columnwidth}{!}{%
\begin{tabular}{llllll}
\toprule
  & BT=0 & BT=5 &BT=10 &BT=15 &BT=20\\
\midrule
\#images       & 4,333  & 4,276 & 3,943 & 3,240 & 2,690     \\
\#food         & 4,033  & 4,012 & 3,790 & 3,207 & 2,704         \\
\#beverage     & 2,239  & 2,235 & 2,145 & 1,875 & 1,651         \\
\bottomrule
\end{tabular}%
}
\caption{Number of image, food, beverage object in training subset with different BTs}
\label{tab:val_freq}
\end{table}

\begin{figure}[t]
    \centering
    \includegraphics[scale = 0.25]{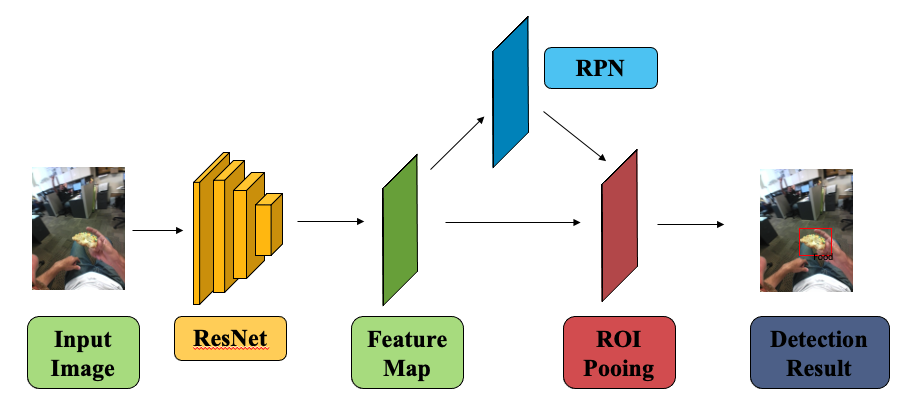}
    \caption{The block diagram of Faster RCNN network}
    \label{fig:rcnn_bd}
\end{figure}

Based on the proposed blur image detection described in Section \emph{Blur Detection}, the only threshold that 
needs to be selected is the Blur Threshold (BT). 
In this section, we describe our experimental design to select  BT. 
\subsection{BT Selection}
The Faster R-CNN was used as our learning-based method to detect food/beverage objects from the AIM captured  images. In order to obtain better recognition results, we adopt transfer learning and used the model pre-trained on ImageNet~\cite{ILSVRC15} as our starting point for training. 
Figure ~\ref{fig:rcnn_bd} shows the structure of our Faster R-CNN network. The ResNet~\cite{10} is used to extract feature maps from the input image, which are then used by the region proposal network (RPN) to identify areas of interest in the image. The ROI pooling layers crop and wrap feature maps using the extracted and generated proposal boxes to obtain fine-tuned box locations and classify the food objects in the image.

As described in Section \emph{Dataset Description}, the dataset is split into training, validation and testing. We vary the BT from 0 to 20 in steps of 5 to create different training datasets. The validation and testing sets remain unchanged. For a given value of BT, all images in the training set with a variance below BT are discarded. As BT increases, the size of our training dataset decreases. We show how the number of images in the training set varies as BT varies in Table~\ref{tab:val_freq}.
Different instances of Faster R-CNN model are trained on each of these training sets for 150 epochs with a batch size of 64. 
The validation set is used to select the model threshold of all Faster R-CNN instances. We use Average Precision (AP) from COCO to evaluate the performance of the object detection model. AP ranges between 0 and 100, with 100 referring to perfect classification. AP is the COCO's standard evaluation metric that averages mean Average Precision (mAP) over different Intersection of Union(IoU) thresholds, from 0.5 to 0.95. More details of AP calculation can be found in \cite{COCO}. 
We report the AP over the validation dataset as BT varies in Table~\ref{tab:val}. The selection of BT is $BT=10$ for our experiments since it gives the best performance on the validation subset.

\subsection{Food Detector Testing}
\begin{figure*}[t!]
	%\centering
	\subfloat[]{\includegraphics[width =4cm]{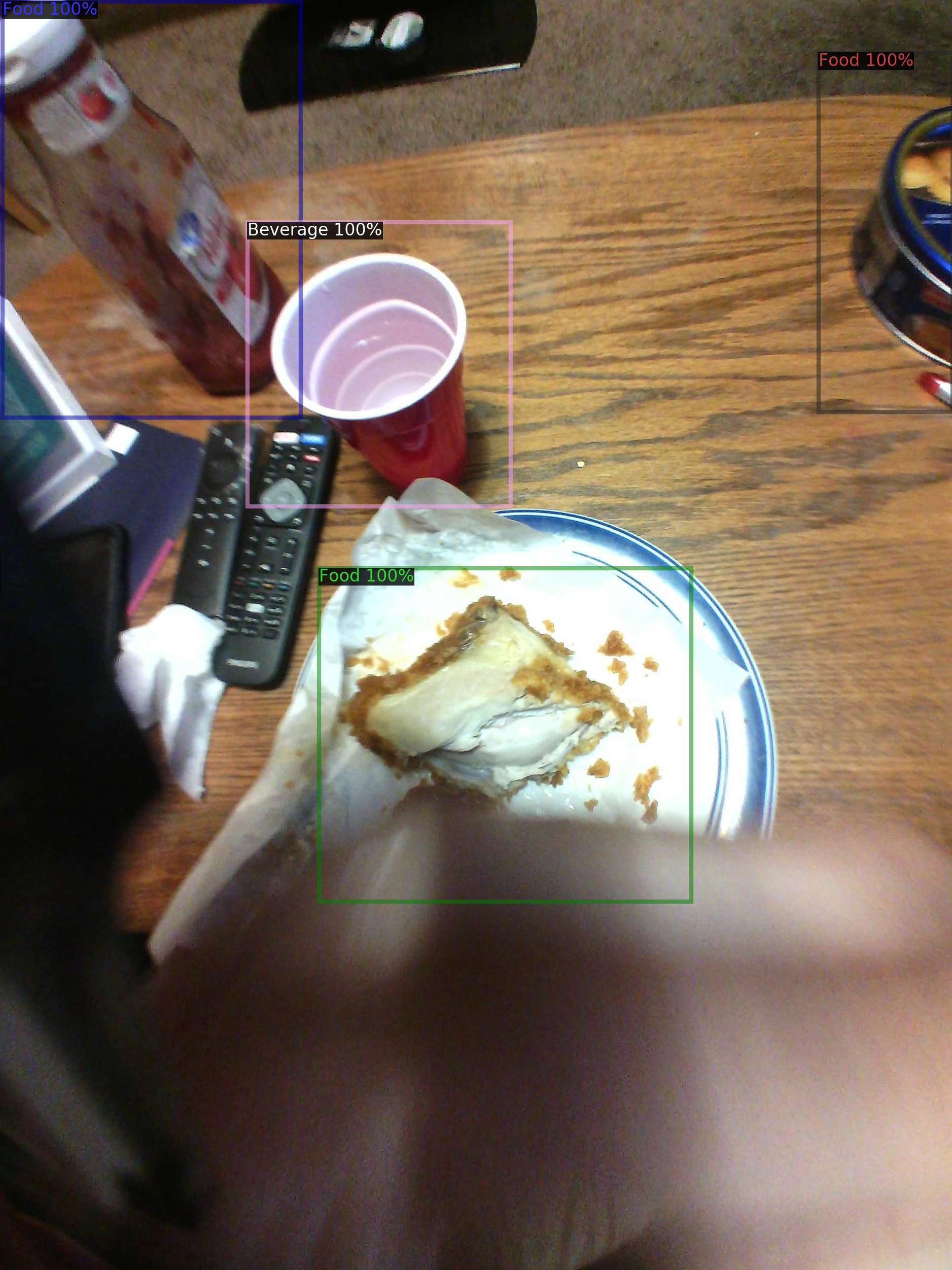}}
	\hfill
	\subfloat[]{\includegraphics[width=4cm]{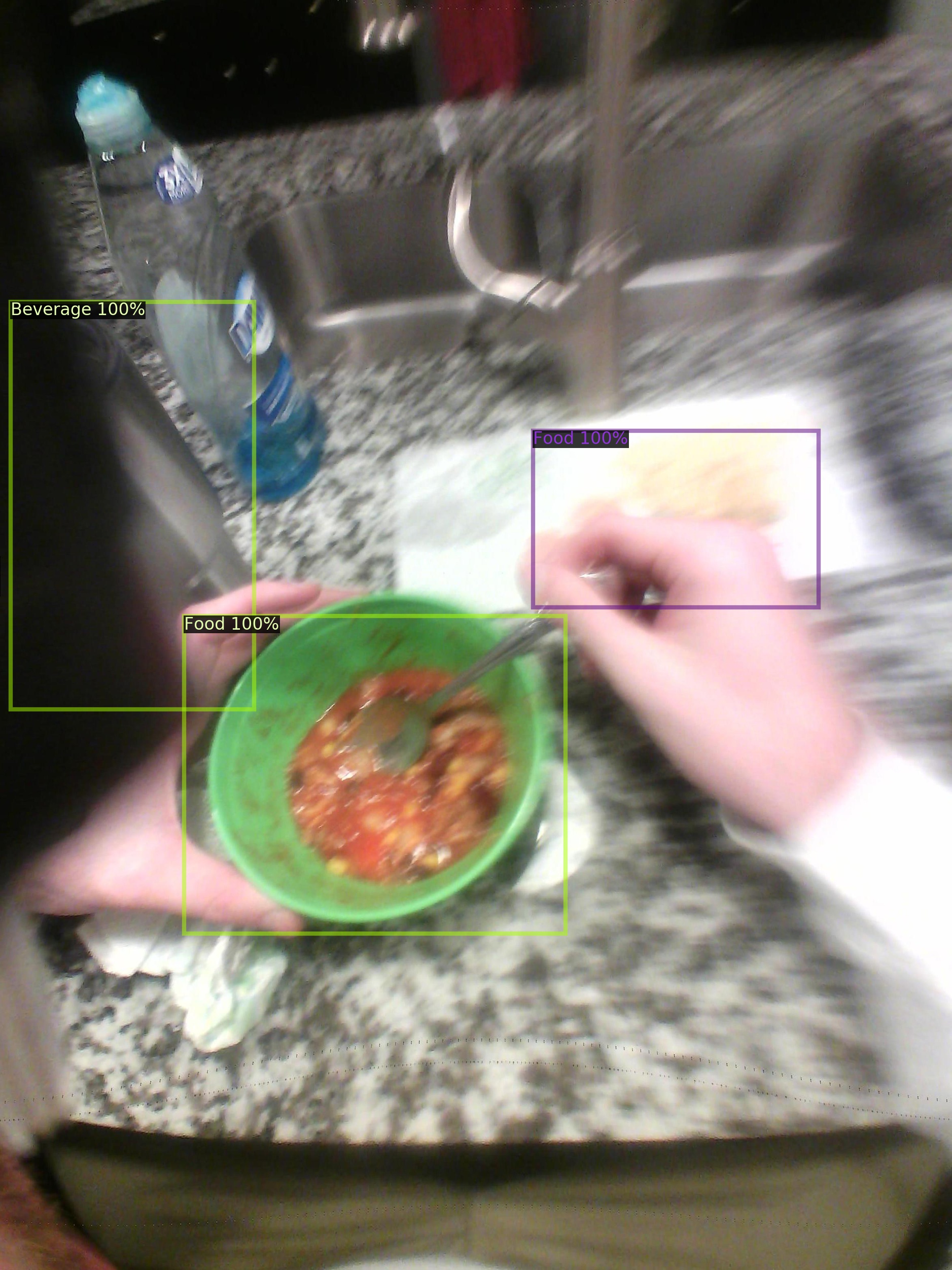}}
	\hfill
	\subfloat[]{\includegraphics[width =4cm]{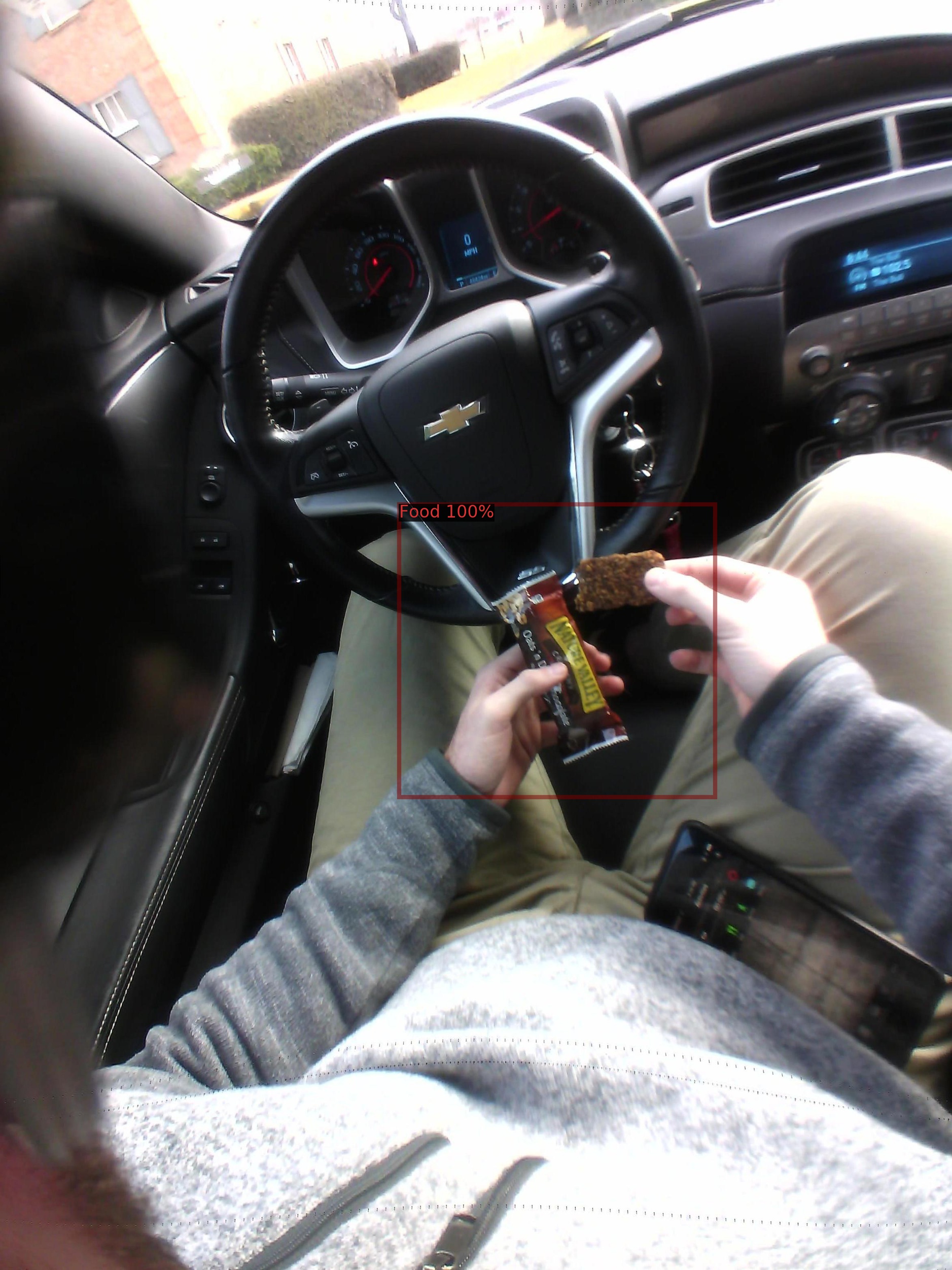}}
	\hfill
	\subfloat[]{\includegraphics[width=4cm]{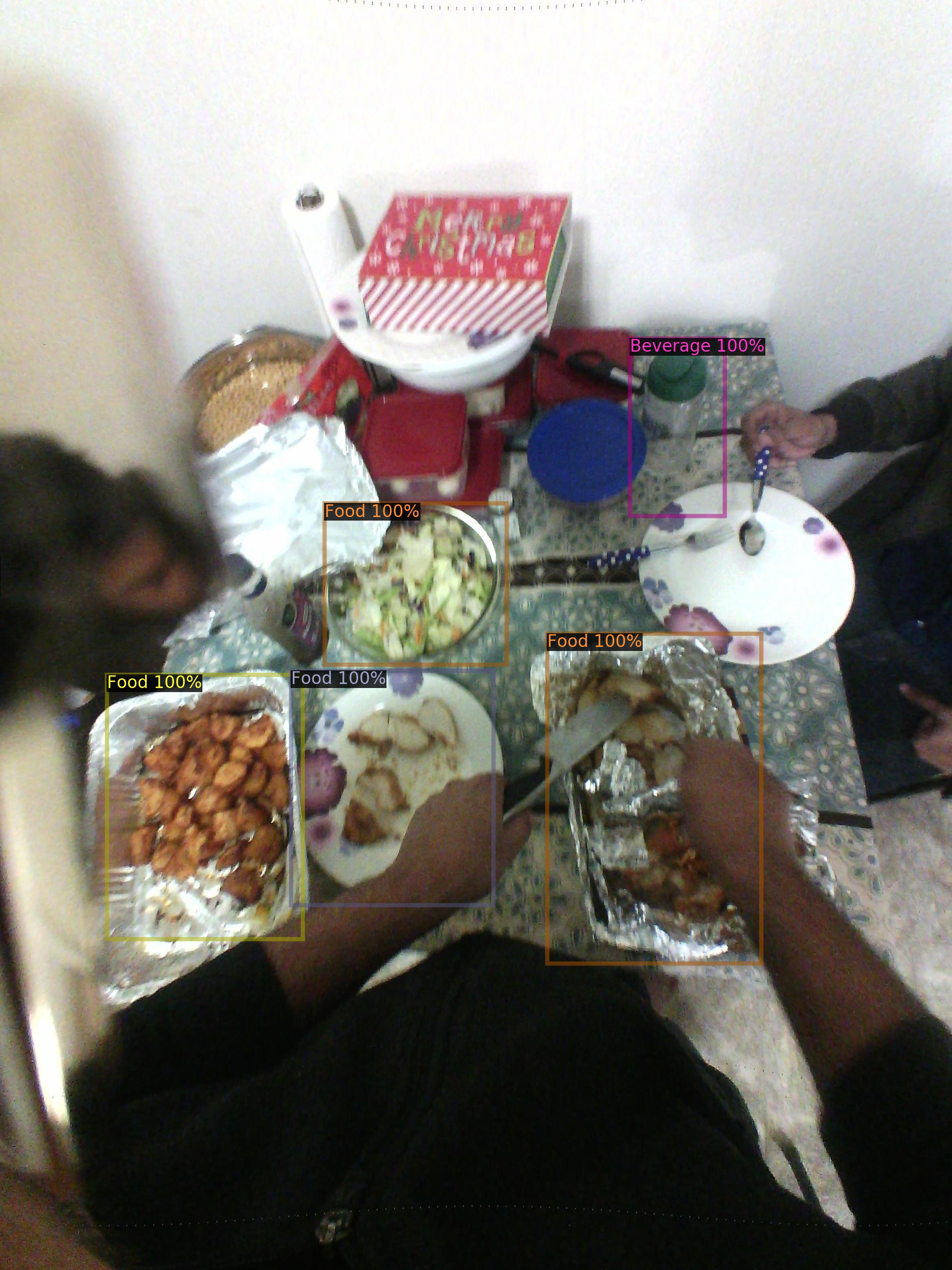}}
	\caption{Object detection results on sample images from the testing subset.}
	\label{fig:test_examples}
\end{figure*}

\begin{table}
\centering
\resizebox{\columnwidth}{!}{%
\begin{tabular}{llllll}
\toprule
  & BT=0 & BT=5 &BT=10 &BT=15 &BT=20\\
\midrule
Overall AP       & 46.37  & 47.07 & \textbf{52.72} & 46.10 & 44.12     \\
AP of food         & 43.47  & 43.95 & \textbf{50.98} & 43.38 & 41.31         \\
AP of beverage     & 49.26  & 50.19 & \textbf{54.45} & 48.82 & 46.92         \\
\bottomrule
\end{tabular}%
}
\caption{Evaluation of detection using the validation subsets with various BT values.}
\label{tab:val}
\end{table}

\begin{table}
\centering
\begin{tabular}{ll}
\toprule
  & BT=10 \\
\midrule
Overall AP       & 51.97     \\
AP of food         & 50.13          \\
AP of beverage     & 53.81          \\
\bottomrule
\end{tabular}
\caption{AP on testing subset with BT = 10.}
\label{tab:test}
\end{table}

From Table~\ref{tab:val} we can see that as BT increases from 0 to 10 the performance of the Faster-RCNN improves across all object categories and as BT increases from 10 to 20 the performance decreases. 
We believe that as BT increases from 0 to 10, our object detection method is seeing a performance increase because it sees fewer extremely blurry images. 
However, as BT increases from 10 to 20 the performance decreases because of a decrease in the number of images in the training dataset. 
This can be verified from Table~\ref{tab:val_freq}. As BT increases from 10 to 15, the training set sees a decrease of 700 images. While BT = 15 removes blurry images from the training set, it does so very aggressively thus removing some images that contain relevant object (food/beverage) features. 
We show AP on the testing set in Table~\ref{tab:test} for the empirically selected value of  BT=10. 
By removing extremely blurry images from the training set, we are able to improve the performance of our object detection system and thus improve the performance of food detection. 
Results of our object detection model on some sample images in the test subset are shown in Figure~\ref{fig:test_examples}. Figure ~\ref{fig:test_examples}(a) and Figure ~\ref{fig:test_examples}(b) shows the detector is still able to locate and classify the item is food/beverage correctly in partially blurry image although some blurry images are excluded from the training set, Figure ~\ref{fig:test_examples}(c) and Figure ~\ref{fig:test_examples}(d) show the detector provides accurate detection results in both scenes with a single item and complex scene with multiple items.

\section{Conclusion}
\label{sec:conclusion}
In this paper, we introduced a food intake sensor, AIM, which captures eating scene images for the purpose of dietary assessment. AIM provides a hands-free automated approach to capture images of eating scene and to provide a precise estimate of the chew count. Images from the AIM device are sometimes affected by blur artifacts which could reduce the performance of various image analysis tasks. We proposed a simple method to improve the food detection performance of images captured by AIM. Experiments were conducted on a dataset consisting of 5,418 eating scene images. We demonstrated that when only extremely blurry images are removed, the performance of the food detection model can be improved. In the future, we plan to further investigate other deblurring methods including machine learning based methods that can be combined for the food analysis task. In addition, we will study how to mitigate the effect of blur on other tasks such as food segmentation, food classification, and portion size estimation.

\bibliographystyle{IEEEbib}
\bibliography{paper}

%%%%%%%%%%%%%%%%%%%%%%%%%%%%%%%%%%
% Biography
%%%%%%%%%%%%%%%%%%%%%%%%%%%%%%%%%%
\begin{biography}
Yue Han received his B.S degree with distinction from Purdue University in 2019. He is currently pursuing a Ph.D. degree at Purdue University and working as a research assistant in the Video and Image Processing Laboratory at Purdue University. His research interests include image processing, computer vision, and deep learning.
\newline

Sri Kalyan Yarlagadda is a machine learning scientist at Overjet. He received his B.Tech in Electrical Engineering from Indian Institute of Technology Madras in July 2015 and his Ph.D in Electrical and Computer Engineering from Purdue University in 2020. His research interests include image processing and computer vision.
\newline

Tonmoy Ghosh received the B.Sc. and M.Sc. degree in Electrical and Electronic Engineering from the Bangladesh University of Engineering and Technology, Dhaka, Bangladesh, in 2012 and 2016, respectively. Currently, he is doing his Ph.D. in Electrical and Computer Engineering at The University of Alabama, Tuscaloosa, USA. His research interests include applying signal processing and machine/deep learning-based method to address engineering problems in images, wearable sensors, with a focus on computer-aided detection and health monitoring applications.
\newline

Fengqing Zhu is an Assistant Professor of Electrical and Computer Engineering at Purdue University, West Lafayette, Indiana. Dr. Zhu received the B.S.E.E. (with highest distinction), M.S. and Ph.D. degrees in Electrical and Computer Engineering from Purdue University in 2004, 2006 and 2011, respectively. Her research interests include image processing and analysis, video compression and computer vision. Prior to joining Purdue in 2015, she was a Staff Researcher at Futurewei Technologies (USA).
\newline

Edward Sazonov received the Ph.D. degree in Computer Engineering from West Virginia University in 2002. Currently he is a James R. Cudworth endowed Professor in the ECE Department at the University of Alabama. His research is focused on development of wearable devices for monitoring of food intake; physical activity and gait; and cigarette smoking. This research was recognized by several awards and supported by the NIH, NSF, NAS, and other agencies, industries and foundations.
\newline

Edward J. Delp is the Charles William Harrison Distinguished Professor
of Electrical and Computer Engineering and Professor of Biomedical
Engineering at Purdue University. His research interests include image
and video processing, image analysis, computer vision, image and
video compression, multimedia security, medical imaging, multimedia
systems, communication and information theory.

\end{biography}

\end{document}